\documentclass{article}

\usepackage[final]{corl_2018} 

\usepackage{times}
\usepackage{relsize}
\usepackage[T1]{fontenc}
\usepackage[scaled=0.90]{inconsolata}
\usepackage[latin1]{inputenc}
\usepackage[english]{babel}

\usepackage{epsfig}
\usepackage{graphicx}
\usepackage{wrapfig}
\usepackage[belowskip=0pt,aboveskip=0pt,font=small]{caption}
\usepackage[belowskip=0pt,aboveskip=0pt,font=small]{subcaption}
\usepackage{amsmath, amsthm, amssymb, mathabx}
\usepackage{textcomp}
\usepackage{stmaryrd}
\SetSymbolFont{stmry}{bold}{U}{stmry}{m}{n}
\usepackage{upgreek}
\usepackage{cases}
\usepackage{mathtools}
\usepackage{arydshln}
\usepackage{multirow}
\usepackage{bbm}

\usepackage{multirow}
\usepackage{rotating}
\usepackage{booktabs}

\usepackage{enumitem}
\usepackage[olditem,oldenum]{paralist}

\usepackage{alltt}
\usepackage{listings}
\usepackage{changepage}

\usepackage{mysymbols}

\usepackage{url}
\usepackage{xspace}
\usepackage{comment}
\usepackage{color}
\usepackage{afterpage}
\usepackage{pdfpages}
\usepackage{framed}
\usepackage{fancybox}
\usepackage{cuted}
\usepackage{microtype}      
\usepackage{nicefrac}       

\usepackage{quoting}
\quotingsetup{vskip=0pt}
\usepackage{epigraph}
\usepackage[dvipsnames]{xcolor}
\usepackage{MnSymbol}

\newcommand{\master}{master\xspace}
\newcommand{\subgoal}{subgoal\xspace}
\newcommand{\subgoals}{subgoals\xspace}

\newcommand{\Subgoals}{Subgoals\xspace}
\newcommand{\subpolicy}{sub-policy\xspace}
\newcommand{\subpolicies}{sub-policies\xspace}
\newcommand{\module}{module\xspace} 
\newcommand{\modules}{modules\xspace} 
\newcommand{\alg}{NMC\xspace}
\newcommand{\algfull}{Neural Modular Controller\xspace}

\newcommand{\subg}[2]{$\langle$\texttt{#1,\emph{#2}}$\rangle$}

\newcommand{\myquote}[1]{\emph{`#1'}}
\newcommand{\myapprox}{{\raise.17ex\hbox{$\scriptstyle\sim$}}}
\newcommand{\xhdr}[1]{\vspace{3pt}\noindent\textbf{#1}}

\newcommand{\reffig}[1]{Fig.~\ref{#1}}
\newcommand{\refsec}[1]{Sec.~\ref{#1}}
\newcommand{\reftab}[1]{Tab.~\ref{#1}}

\newcommand{\eqa}{EmbodiedQA\xspace}
\newcommand{\eqads}{EQA\xspace}

\title{Neural Modular Control for\\ Embodied Question Answering}

\author{
    Abhishek Das$^{1^\star}$ \hspace{0.5pc}
    Georgia Gkioxari$^2$ \hspace{0.5pc}
    Stefan Lee$^1$ \hspace{0.5pc}
    Devi Parikh$^{1,2}$ \hspace{0.5pc}
    Dhruv Batra$^{1,2}$ \\[0.05in]
    {\small $^1$Georgia Institute of Technology} \hspace{0.5pc}
    {\small $^2$Facebook AI Research}
}

\begin{document}

\maketitle
\renewcommand*{\thefootnote}{$\star$}
\setcounter{footnote}{1}
\footnotetext{Work partially done during an internship at Facebook AI Research.}
\renewcommand*{\thefootnote}{\arabic{footnote}}
\setcounter{footnote}{0}
\thispagestyle{empty}

\begin{abstract}

We present a modular approach for learning policies for navigation
    over long planning horizons from language input. Our hierarchical policy
    operates at multiple timescales, where the higher-level master policy proposes
    subgoals to be executed by specialized \subpolicies. Our choice of subgoals
    is compositional and semantic, \ie they can be sequentially combined in
    arbitrary orderings, and assume human-interpretable descriptions (\eg `exit room',
    `find kitchen', `find refrigerator', \etc.).

We use imitation learning to warm-start policies at each level of the hierarchy,
    dramatically increasing sample efficiency, followed by reinforcement learning.
    Independent reinforcement learning at each level of hierarchy enables \subpolicies
    to adapt to consequences of their actions and recover from errors. Subsequent
    joint hierarchical training enables the master policy to adapt to the \subpolicies.

On the challenging EQA~\cite{embodiedqa} benchmark in House3D~\cite{house3d},
    requiring navigating diverse realistic indoor environments, our approach
    outperforms prior work by a significant margin, both in terms of navigation
    and question answering.

\end{abstract}


\section{Introduction}
\label{sec:intro}

Abstraction is an essential tool for navigating our daily lives.
When seeking a late night snack, we certainly do not spend time
planning out the mechanics of walking and are thankfully also
unburdened of the effort of recalling to beat our heart along the
way. Instead, we conceptualize our actions as a series of higher-level semantic goals
-- {\small\texttt{exit bedroom; go to kitchen; open fridge; find snack;}}
-- each of which is executed through specialized coordination of
our perceptual and sensorimotor skills. This
ability to abstract long, complex sequences of actions into semantically
meaningful \subgoals is a key component of human cognition \cite{roth_cogsci79} and it is natural to
believe that artificial agents can benefit from applying similar mechanisms when navigating our world.

We study such hierarchical control in the context of a recently proposed task --
Embodied Question Answering (EmbodiedQA)~\cite{embodiedqa} -- where an embodied agent is spawned
at a random location in a novel environment (\eg a house) and asked to answer a question
(\myquote{What color is the piano in the living room?}).
To do so, the agent must navigate from egocentric vision alone (without access
to a map of the environment), locate the entity in question (\myquote{piano in the living room}),
and respond with the correct answer (\eg \myquote{red}). From a reinforcement learning (RL) perspective, EmbodiedQA
presents challenges that are known to make learning particularly difficult --
partial observability, planning over long time horizons, and sparse rewards -- the
agent may have to navigate through multiple rooms in search for the
answer, executing hundreds of primitive motion actions
along the way ({\small\texttt{forward; forward; turn-right;}} \ldots) and receiving a reward based only on its final answer.

\begin{figure}[t]
  \centering
  \includegraphics[width=\linewidth]{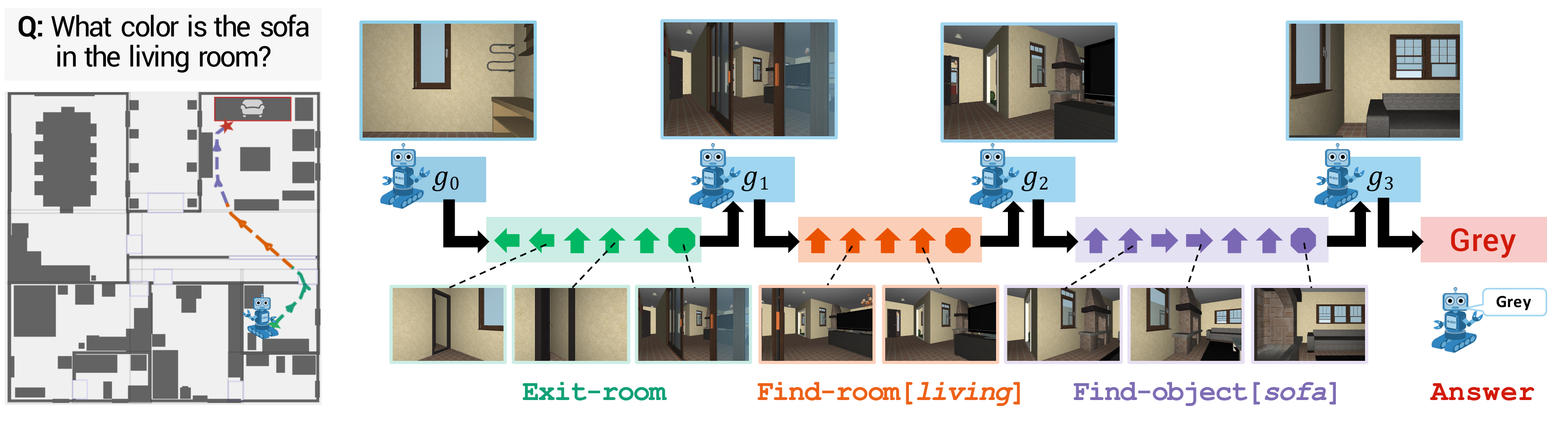}
  \caption{We introduce a hierarchical policy for Embodied Question Answering.
  Given a question (``What color is the sofa in the living room?'') and observation,
  our master policy predicts a sequence of subgoals -- {\small\texttt{Exit-room}},
  {\small\texttt{Find-room[living]}}, {\small\texttt{Find-object[sofa]}}, {\small\texttt{Answer}}
  -- that are then executed by specialized \subpolicies to navigate to the target
  object and answer the question (``Grey'').}
  \label{fig:teaser}
\end{figure}

To address this challenging learning problem, we develop a hierarchical \textbf{\algfull (\alg)}~--
consisting of a \emph{\master} policy that determines high-level \emph{\subgoals},
and \emph{\subpolicies} that execute a series of low-level actions to achieve these \subgoals.
Our \alg model constructs a hierarchy that is arguably natural to this problem --
navigation to rooms and objects \vs low-level motion actions.
For example, \alg seeks to break down
a question \myquote{What color is the piano in the living room?} to the series of \subgoals
{\small\texttt{ exit-room; find-room[\emph{living}]; find-object[\emph{piano}]; answer;}}
and execute this plan with specialized neural `\modules'
corresponding to each \subgoal. Each \module is trained to issue
a variable length series of primitive actions to achieve its titular \subgoal~-- \eg the {\small\texttt{find-object[\emph{piano}]}}
module is trained to navigate the agent to the input argument \emph{piano} within the current room.
Disentangling semantic \subgoal selection from \subpolicy execution results in
easier to train models due to shorter time horizons. Specifically, this hierarchical structure introduces:\\[-11pt]
\begin{compactenum}[\hspace{0pt}--]

\item \textbf{Compressed Time Horizons:}
The \master policy makes orders of magnitude fewer decisions over the course of a
navigation than a \emph{`flat model'} that directly predicts primitive actions -- allowing
the answering reward in \eqa to more easily influence high-level motor control decisions.

\item \textbf{Modular Pretraining:}
As each \module corresponds to a specific task, they can be trained independently before
being combined with the \master policy. Likewise, the \master policy can be trained
assuming ideal \modules. We do this through imitation learning~\cite{abbeel_icml04,ross_arxiv14}
\subpolicies.

\item \textbf{Interpretability:} The predictions made by the \master policy correspond to
semantic \subgoals and exposes the reasoning of the agent to inspection
(\myquote{What is the agent trying to do right now?}) in a significantly more interpretable
fashion than just its primitive actions. 

\end{compactenum}

First, we learn and evaluate master and \subpolicies for each of our \subgoals, trained using
behavior cloning on expert trajectories, reinforcement learning from scratch, and
reinforcement learning after behavior cloning. We find that reinforcement learning
after behavior cloning dramatically improves performance over each individual training regime.
We then evaluate our combined hierarchical approach on the EQA~\cite{embodiedqa} benchmark in House3D~\cite{house3d} environments.
Our approach significantly outperforms prior work both in navigational and question answering performance --
our agent is able to navigate closer to the target object
and is able to answer questions correctly more often.
\section{Related Work}
\label{sec:related}

Our work builds on and is related to prior work in hierarchical reinforcement and imitation learning,
grounded language learning, and embodied question-answering agents in simulated environments.

\textbf{Hierarchical Reinforcement and Imitation Learning.}
Our formulation is closely related to Le~\etal~\cite{le_icml18},
and can be seen as an instantiation of the options framework~\cite{sutton_ai99,sutton_icml98},
wherein a global \master policy proposes \subgoals~-- to be achieved by local \subpolicies~--
towards a downstream task objective~\cite{bacon_aaai17,kulkarni_nips16,tessler_aaai17}.
Relative to other work on automatic \subgoal discovery in hierarchical
reinforcement learning~\cite{bakker_ias04,goel_aaai03,mcgovern_icml01},
we show that given knowledge of the problem structure, simple heuristics are quite
effective in breaking down long-range planning into sequential \subgoals.
We make use of a combination of hierarchical behavior cloning~\cite{abbeel_icml04}
and actor-critic~\cite{mnih_icml16} to train our modular policy.

\textbf{Neural Module Networks and Policy Sketches.}
At a conceptual-level, our work is analogous to recent work on
neural module networks (NMNs)~\cite{andreas_icml17,andreas_naacl16,andreas_cvpr16}
for visual question answering. NMNs first predict a `program' from the question,
consisting of a sequence of primitive reasoning steps, which are then executed on the image to
obtain the answer.
Unlike NMNs, where each primitive reasoning module has access to the entire image (completely observable)
our setting is partially observable -- each \subpolicy only has access to first-person RGB --
making active re-evaluation of \subgoals after executing each \subpolicy essential.
Our work is also closely related to policy sketches \cite{andreas_icml17}, which
are symbolic descriptions of \subgoals provided to the agent without any grounding or \subpolicy for executing them.
There are two key differences \wrt to our work.
First, an important framework difference -- Andreas \etal \cite{andreas_icml17} assume
access to a policy sketch \emph{at test time}, \ie for every task to be performed. %
In EmbodiedQA, this would correspond
to the agent being provided with a high-level plan ({\small\texttt{exit-room; find-room[\emph{living}];}} ...) for every question
it is ever asked, which is an unrealistic assumption in real-world scenarios with a robot. In contrast, we assume that
\subgoal supervision (in the form of expert demonstrations and plans)
are available on training environments but not on test,
and the agent must \emph{learn} to produce its own \subgoals. Second, a subtle but important implementation difference --
unlike \cite{andreas_icml17}, our \subpolicy modules accept input arguments that are embeddings of target rooms and objects
(\eg {\small\texttt{find-room[\emph{living}], find-object[\emph{piano}]}}).
This results in our \subpolicy modules being shared not just across tasks (questions) as in \cite{andreas_icml17}, but also
across instantiations of \emph{similar} navigation \subpolicies~-- \ie, {\small\texttt{find-object[\emph{piano}]}} and
{\small\texttt{find-object[\emph{chair}]}} share parameters that enable data efficient learning
without exhaustively learning separate policies for each.


\textbf{Grounded Language Learning.}
Beginning with SHRDLU~\cite{winograd_cogpsy72}, there has been a rich progression
of work in grounding language-based goal specifications into actions and pixels
in physically-simulated environments.
Recent deep reinforcement learning-based approaches to this
explore it in 2D gridworlds~\cite{andreas_icml17,yu_iclr18,misra_acl17},
simple visual~\cite{chaplot_aaai18,hermann_arxiv17,hill_arxiv17,oh_icml17,shu_iclr18,vogel_acl10}
and textual~\cite{matuszek_iser13,narasimhan_emnlp15} environments,
perceptually-realistic 3D home simulators~\cite{embodiedqa,gordon_cvpr18,puig_cvpr18,zhu_iccv17,zhu_icra17},
as well as real indoor scenes~\cite{anderson_cvpr18,gupta_cvpr17,gupta_arxiv17}.
Our hierarchical policy learns to ground words from the question into
two levels of hierarchical semantics. The master policy grounds words into \subgoals
(such as {\small\texttt{find-room[\emph{kitchen}]}}), and \subpolicies ground these semantic targets
(such as {\small\texttt{\emph{cutting board}}}, {\small\texttt{\emph{bathroom}}}) into primitive actions and raw pixels,
both parameterized as neural control policies and trained end-to-end.

\textbf{Embodied Question-Answering Agents.} Finally, hierarchical policies for
    embodied question answering have previously been proposed by
    Das~\etal~\cite{embodiedqa} in the House3D environment~\cite{house3d}, and by
    Gordon~\etal~\cite{gordon_cvpr18} in the AI2-THOR environment~\cite{kolve_arxiv17}.
    Our hierarchical policy, in comparison, is human-interpretable,
    \ie the \subgoal being pursued at every step of navigation is semantic,
    and due to the modular structure, can navigate over longer paths than prior work,
    spanning multiple rooms.

\section{Neural Modular Control}
\label{sec:approach}

We now describe our approach in detail.
Recall that given a question, the goal of our agent is to predict
a sequence of navigation \subgoals and execute them to ultimately find the target object
and respond with the correct answer.
We first present our modular hierarchical policy.
We then describe how we extract optimal plans from shortest
path navigation trajectories for behavior cloning. And finally, we describe how
the various modules are combined and trained with a
combination of imitation learning (behavior cloning) and reinforcement learning.

\subsection{Hierarchical Policy}
\label{sec:policy}

\xhdr{Notation.}
Recall that NMC has 2 levels in the hierarchy -- a \master policy that generates \subgoals and
\subpolicies for each of these \subgoals. We use $i$ to index the sequence of \subgoals
and $t$ to index actions generated by \subpolicies.
Let $\mathcal{S} = \{s\}$ denote the set of states, $\mathcal{G} = \{g\}$ the set of variable-time
\subgoals with elements $g= \langle g_{\text{task}}, g_{\text{argument}} \rangle$,
 \eg $g = \langle$\texttt{exit-room,\emph{None}}$\rangle$, or $g = \langle$\texttt{find-room,\emph{bedroom}}$\rangle$.
Let $\mathcal{A} = \{a\}$ be the set of primitive actions (\texttt{forward}, \texttt{turn-left}, \texttt{turn-right}).
The learning problem can then be succinctly put as learning a master policy $\pi_\theta: \mathcal{S} \rightarrow \mathcal{G}$
parameterized by $\theta$ and \subpolicies $\pi_{\phi_g}: \mathcal{S} \rightarrow \mathcal{A} \cup \{\text{\texttt{stop}}\}$
parameterized by $\phi_g, \, \forall g \in \mathcal{G}$, where
the \texttt{stop} action terminates a \subpolicy and returns
control to the \master policy.

While navigating an environment, control alternates between the master policy selecting \subgoals and \subpolicies
executing these goals through a series of primitive actions. More formally, given an initial state $s_0$ the master
policy predicts a subgoal $g_0 \sim \pi_\theta(g \vert s_0)$, the corresponding \subpolicy executes until
some time ${T}_{0}$
when either (1) the \subpolicy terminates itself by producing the \texttt{stop} token
$a_{T_0} \sim \pi_{\phi_{g_0}}(a \vert s_{T_0}) = \text{\texttt{stop}}$ or (2) a maximum number of primitive actions has been reached.
Either way, this returns the control back to the master policy
which predicts another \subgoal and repeats this process until termination. This results in a
state-\subgoal trajectory:
\begin{align}
\Sigma = \bigg(\underbrace{s_0, g_0}_{\text{\subgoal 0}}, \underbrace{s_{T_0}, g_{1}}_\text{\subgoal 1},
\ldots, \underbrace{s_{T_{i}}, g_{i+1}}_{\text{\subgoal} \,\, i},
\ldots, \underbrace{s_{T_{\mathcal{T}-1}}, g_\mathcal{T}}_{\text{\subgoal } \,\, \mathcal{T}}\bigg)
\end{align}
for the master policy. Notice that the terminal state of the $i^{\text{th}}$ \subpolicy
$s_{T_i}$ forms the state for the \master policy to predict the next \subgoal $g_{i+1}$.
For the $(i+1)^\text{th}$ \subgoal $g_{i+1}$, the low-level trajectory of states and primitive actions is given by:
\begin{align}
\sigma_{g_{i+1}} = \bigg( \underbrace{s_{T_{i}}, a_{T_{i}}}_{\text{action 0}},
\underbrace{s_{T_{i}+1}, a_{T_{i}+1}}_{\text{action 1}}, \ldots,
\underbrace{s_{T_{i}+t}, a_{T_{i}+t}}_{\text{action t}}, \ldots, s_{T_{i+1}}
\bigg).
\end{align}
%
Note that by concatenating all \subpolicy trajectories in order
$(\sigma_{g_0}, \sigma_{g_1}, \ldots, \sigma_{g_\calT})$,
the entire trajectory of states and primitive actions can be recovered.

\xhdr{\Subgoals $\langle\text{Tasks, Arguments}\rangle$.}
As mentioned above, each \subgoal is factorized into a task and an argument
$g= \langle g_{\text{task}}, g_{\text{argument}}\rangle$. There are 4 possible tasks --
\texttt{exit-room}, \texttt{find-room}, \texttt{find-object}, and \texttt{answer}.
Tasks \texttt{find-object} and \texttt{find-room}
accept as arguments one of the 50 objects and 12 room types in
EQA v1 dataset \cite{embodiedqa} respectively;
\texttt{exit-room} and \texttt{answer} do not accept any arguments. This gives us a total of $50+12+1+1 = 64$
\subgoals.
\begin{adjustwidth}{0.2in}{0in}
	\subg{exit-room}{none},  \hspace{1.815in} \subg{answer}{none}, \hspace{0.2in} \big\} 0 args\\
	\subg{find-object}{couch}, \subg{find-object} {cup}, \ldots , \subg{find-object}{xbox}, \hspace{0.2in} \big\} 50 args\\
	\subg{find-room}{living}, \subg{find-room}{bedroom}, \ldots , \subg{find-room}{patio}. \hspace{0.185in} \big\} 12 args
\end{adjustwidth}

Descriptions of these tasks and their success criteria are provided in \tableref{tab:heuristics}.
\begin{table*}
    \begin{center}
    \resizebox{0.95\columnwidth}{!}{
    \begin{tabular}{p{2cm}p{2.1cm}p{4.3cm}p{3.05cm}}
    \toprule
    \textbf{Subgoal} & \textbf{Argument(s)} & \textbf{Description} & \textbf{Success} \\
    \toprule
    	\small{\texttt{Exit-room}}
        & \small{None}
        & \small{When there is only 1 door in spawn room, or 1 door other than door entered through
            in an intermediate room; agent is forced to use the remaining door.}
        & \small{Stopping after exiting through the correct door.} \\ 
\midrule
    	\vspace{-15pt}\small{\texttt{Find-room}}
        & \small{\shortstack[l]{Room name\\ (\textit{gym,  kitchen, ...})}}
        & \vspace{-15pt}\small{When there are multiple doors and the agent has to search
            and pick the door to the target room.}
        & \vspace{-15pt}\small{Stopping after entering target room.} \\ 
\midrule
    	\vspace{-15pt}\small{\texttt{Find-object}}
        & \small{\shortstack[l]{Object name\\ (\textit{oven,  sofa, ...})}}
        & \vspace{-15pt}\small{When the agent has to search for a specific object in room.}
        & \vspace{-15pt}\small{Stopping within $0.75\text{m}$ of the target object.} \\ 
\midrule
    	\small{\texttt{Answer}}
        & \small{None}
        & \small{When the agent has to provide an answer from the answer space.}
        & \small{Generating the correct answer to the question.} \\ 
    \bottomrule
    \end{tabular}}\\[3pt]
    \caption{Descriptions of our subgoals and conditions we use to extract
        them automatically from expert trajectories.}
    \label{tab:heuristics}
    \end{center}
\end{table*}

\xhdr{Master Policy.} The master policy $\mathbf{\pi_\theta}$ parameterized
by $\theta$ is implemented as a single layer Gated
Recurrent Unit (GRU). At each high-level step $i+1$, the master policy $\mathbf{\pi_\theta}(g \vert s_{T_i})$
takes as input the concatenation of a encoding of the question $q\in\mathbb{R}^{128}$,
the image feature $v_{T_{i}}\in\mathbb{R}^{128}$ of the current frame and an
encoding $o_{i}\in\mathbb{R}^{32}$ computed from a 1-hot representation of the $i^{\text{th}}$
\subgoal, \ie $\mathbbm{1}(g_{i})$.
This information is used to update the hidden state $h_i \in \mathbb{R}^{1048}$
that encodes the entire trajectory up to time $t$ and serves as the state representation.
The policy then produces a probability
distribution over all possible (64) \subgoals $\mathcal{G}$.
We train these policies with actor-critic methods and thus the network also produces
a value estimate. 

\xhdr{Sub-policies.}
%
To take advantage of the comparatively lower number of subgoal tasks, we
decompose \subpolicy parameters $\phi_g$ into
$\phi_{g_\text{task}}$ and $\phi_{g_\text{argument}}$, where
$\phi_{g_\text{task}}$ are shared across the same task and
$\phi_{g_\text{argument}}$ is an argument specific embedding.
Parameter sharing enables us to learn the shared task
in a sample-efficient manner, rather than exhaustively
learning separate \subpolicies for each combination.

Like the master policy, each \subpolicy $\pi_{\phi_{g}}$ is implemented as a
single-layer GRU. At each low-level time step $t$, a \subpolicy $\pi_{\phi_{g}}(a \vert s_t)$ takes as input
the concatenation of the image feature $v_{t}\in\mathbb{R}^{128}$
of the current frame, an encoding $p_{t-1}\in\mathbb{R}^{32}$
computed from a 1-hot representation of the
previous primitive action \ie $\mathbbm{1}(a_{t-1})$,
and the argument embedding $\phi_{g_\text{argument}}$. These inputs are used to update the
hidden state $h_t^g\in\mathbb{R}^{1048}$ which serves as the
state representation. The policy then outputs a
distribution over primitive actions (\texttt{forward, turn-left,
turn-right, stop}). As with the master policy,
each \subpolicy also output a value estimate. 
shows this model structure.

\xhdr{Perception and Question Answering.} To ensure fair comparisons to prior work,
we use the same perception and question answering models as used by Das~\etal~\cite{embodiedqa}.
The perception model is a simple convolutional neural network trained
to perform auto-encoding, semantic segmentation, and depth estimation
from RGB frames taken from House3D \cite{house3d}. Like \cite{embodiedqa},
we use the bottleneck layer of this model as a fixed feature extractor.
We also use the same post-navigational question-answering model as \cite{embodiedqa}, which encodes the question
with a 2-layer LSTM and performs dot-product based attention between the question
encoding and the image features from the last five frames along the navigation path right before the \texttt{answer}
module is called. This post-navigational answering module is trained using visual features along the
shortest path trajectories and then frozen. By keeping these parts of the architecture identical to \cite{embodiedqa},
our experimental comparisons can focus on the differences only due to our contributions, the
\algfull.

\vspace{-5pt}
\subsection{Hierarchical Behavior Cloning from Expert Trajectories}
\vspace{-5pt}

The questions in EQA v1 dataset \cite{embodiedqa} (\eg \myquote{What color is the fireplace?})
are constructed to inquire about attributes (color, location, \etc) of specific target objects (\myquote{fireplace}).
This notion of a target enables the construction of an automatically generated
\emph{expert trajectory} $(s_0^*, a_0^*, \ldots, s_T^*, a_T^*)$ --
the states and actions along the shortest path from the agent spawn
location to the object of interest specified in the question. Notice that these shortest paths
may only be used as supervision on training environments 
but may not be utilized during evaluation on test environments
(where the agent must operate from egocentric vision alone).

Specifically, we would like to use these expert demonstrations to pre-train our proposed \alg navigator
using behavior cloning. However, these trajectories $(s_0^*, a_0^*, \ldots, s_T^*, a_T^*)$ correspond to
a series of primitive actions. To provide supervision for 
both the master policy and \subpolicies, these shortest-path trajectories must be
annotated with a sequence of \subgoals and segmented into their respective temporal extents,
resulting in $\Sigma^*$ and $(\sigma_{g_i}^*)$.

\begin{figure}[h]
    \begin{subfigure}[b]{0.43\textwidth}
        \centering
        \includegraphics[width=\textwidth]{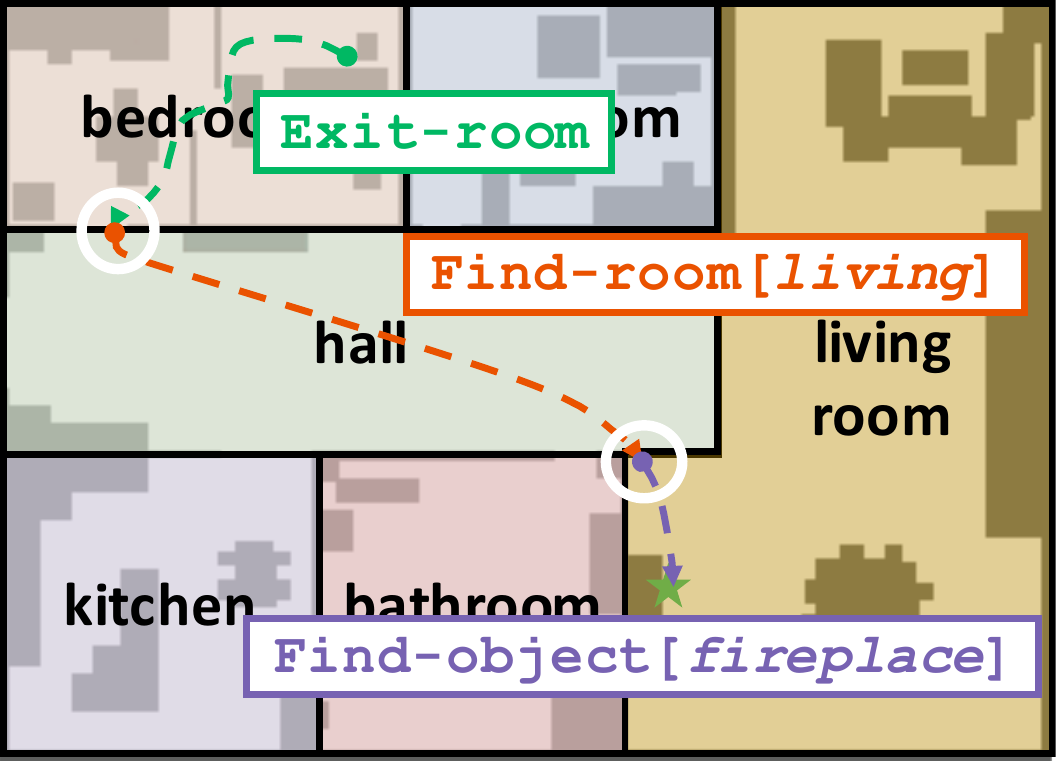}\\[5pt]
        \caption{\textbf{Q}: What color is the fireplace? \textbf{A}: Brown}
        \label{fig:plan_example}
    \end{subfigure}
    \hfill
    \begin{subfigure}[b]{0.52\textwidth}
        \centering
        \includegraphics[width=\textwidth]{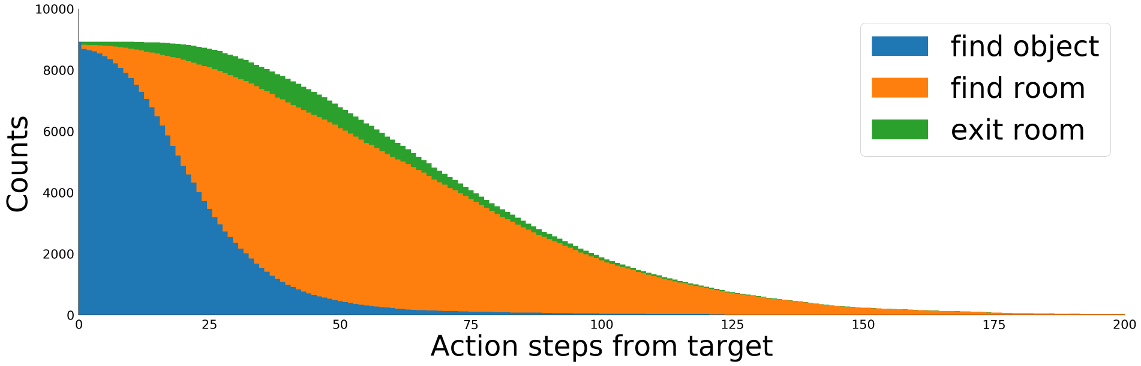}\\[5pt]
        \caption{Distribution of subgoals with number of actions from
        the target object as per expert plans. Closer to the target object,
        the expert plan predominantly consists of \small{\texttt{Find-object}}, while as we move
        farther away, the proportion of \small{\texttt{Find-room}} and \small\texttt{Exit-room} goes up.}
        \label{fig:plan_dist}
    \end{subfigure}
    \\[2pt]
    \caption{We extract expert subgoal trajectories from shortest paths by dividing paths on room transition boundaries (circled in (a))and following the rules in \reftab{tab:heuristics}.}
\end{figure}

We automate this `lifting' of annotation up the hierarchy by
leveraging the object and room bounding boxes provided by the House3D \cite{house3d}.
Essentially, a floor plan may be viewed as an undirected graph with rooms as nodes and doorways
as edges connecting a pair of adjacent rooms.
An example trajectory is shown in \figref{fig:plan_example}
for the question \myquote{What color is the fireplace?}. The agent is spawned in a bedroom,
the shortest path exits into the hall, enters the living room, and approaches the fireplace.
We convert this trajectory to the subgoal sequence (\texttt{exit-room, find-room[\emph{living}],
find-object[\emph{fireplace}], answer}) by recording the transitions on the shortest path from one room
to another, which also naturally provides us with temporal extents of these \subgoals.

We follow a couple of subtle but natural rules: (1) \texttt{find-object} is tagged only when the agent has reached the
destination room containing the target object;
and (2) \texttt{exit-room} is tagged only when the `out-degree' of the current room in the floor-plan-graph
is exactly 1 (\ie either the current room has exactly one doorway or the current room has two
doorways but the agent came in through one).
Rule (2) ensures a semantic difference between \texttt{exit-room} and \texttt{find-room} --
informally, \texttt{exit-room} means \myquote{get me out of here}
and \texttt{find-room[\emph{name}]} means \myquote{look for room \texttt{name}}.

\reftab{tab:heuristics} summarizes these \subgoals and the heuristics used to
    automatically extract them from navigational paths. \reffig{fig:plan_dist} shows
    the proportions of these \subgoals in expert trajectories as a function of the distance
    from target object. Notice that when the agent is close to the target, it is likely to be within the same room
    as the target and thus \texttt{find-object} dominates. On the other hand,
    when the agent is far away from the target, \texttt{find-room}
    and \texttt{exit-room} dominate.

We perform this lifting of shortest paths for all training set questions in
EQA v1 dataset \cite{embodiedqa}, resulting in $N$ expert trajectories
$\{\Sigma_n^*\}_{n=1}^N$ for the \master policy
and $K (>>N)$ trajectories $\{\sigma_{g_k}^* \}_{k=1}^K$ for \subpolicies.
We can then perform hierarchical behavior cloning by minimizing the sum of
cross-entropy losses over all decisions in all expert trajectories.
As is typical in maximum-likelihood training of directed probabilistic models
(\eg hierarchical Bayes Nets), full supervision results in decomposition
into independent sub-problems.
Specifically, with a slight abuse of notation, let $(s_i^*, g_{i+1}^*) \in \Sigma^*$ denote an iterator
over all state-\subgoal tuples in $\Sigma^*$, and {\small$\displaystyle \sum_{(s_i^*, g_{i+1}^*) \in \Sigma^*}$}
denote a sum over such tuples.

Now, the independent learning problems can be written as:
{\small
\begin{subequations}
\begin{align}
\theta^* = \argmin_\theta \,\, & \sum_{n=1}^{N}  \hspace{0.67in}
\sum_{ (s_{i}^*, g_{i+1}^*) \in \Sigma_n^*} \,\,\,\,
-\log\Big(\pi_\theta(g_{i+1}^* \vert s_{i}^*)\Big)
& \text{(\master policy cloning)} \label{eq:master_cloning} \\
\phi_g^* = \argmin_\phi \,\, &
\underbrace{   \vphantom{ \sum_{ (s_{t}^*, a_{t+1}^*) \in \sigma_{g_k}^*}}
\sum_{k=1}^{K} \, \ind{g_k = g}}_{\text{demonstrations}} \,\,\,\,\,
\underbrace{\sum_{ (s_{t}^*, a_{t+1}^*) \in \sigma_{g_k}^*}}_{\text{transitions}} \,\,\,
\underbrace{ \vphantom{ \sum_{ (s_{t}^*, a_{t+1}^*) \in \sigma_{g_k}^*}}
-\log\Big(\pi_{\phi_g}(a_{t+1}^* \vert s_{t}^*)\Big)}_{\text{negative-log-likelihood}}
& \text{(\subpolicy cloning)} \label{eq:subpolicy_cloning}
\end{align}
\end{subequations}
}

Intuitively, each \subpolicy independently maximizes the conditional probability of actions observed in the
expert demonstrations, and the \master policy essentially trains assuming perfect \subpolicies.

\subsection{Asynchronous Advantage Actor-Critic (A3C) Training}
\label{sec:a3c}

After the independent behavior cloning stage, the policies have learned to mimic expert
trajectories; however, they have not had to coordinate with each other or recover from
their own navigational errors. As such, we fine-tune them with
reinforcement learning -- first independently and then jointly.

\textbf{Reward Structure.} The ultimate goal of our agent is to answer questions accurately; however, doing so
requires navigating the environment sufficiently well in search of the answer. We
mirror this structure in our reward $R$, decomposing it into a sum of a sparse
terminal reward $R_{\text{terminal}}$ for the final outcome and a dense, shaped
reward $R_\text{shaped}$ \cite{ng_icml99} determined by the agent's progress towards its goals.
For the master policy $\pi_\theta$, we set $R_\text{terminal}$ to be 1 if
the model answers the question correctly and 0 otherwise. The shaped
reward $R_\text{shaped}$ at \master-step $i$ is based on the change of
navigable distance to the target object before and after executing \subgoal $g_i$.
Each \subpolicy $\pi_{\phi_g}$ also has a terminal 0/1 reward $R_\text{terminal}$
for stopping in a successful state, \eg~\texttt{Exit-room} ending
outside the room it was called in (see \reftab{tab:heuristics} for
all success definitions). Like the master policy,  $R_\text{shaped}$ at
time $t$ is set according to the change in navigable distance to the \subpolicy
target (\eg a point just inside a living room for
\texttt{find-room[\emph{living}]}) after executing
the primitive action $a_t$.  Further, \subpolicies are also
penalized a small constant (-0.02) for colliding with obstacles.

\textbf{Policy Optimization.}
We update the master and \subpolicies to maximize expected
discounted future rewards $J(\pi_\theta)$ and $J(\pi_{\phi_g})$
respectively through the Asynchronous Advantage Actor
Critic \cite{mnih_icml16} policy-gradient algorithm.
Specifically, for the master policy, the gradient of the expected reward is
written as:
 %
 \begin{equation}
 \nabla_\theta J(\pi_\theta) = \mathbb{E}\left[ \nabla_\theta \log(\pi_\theta(g_i | s_{T_i})) \left(Q(s_{T_i}, g_i) - c_\theta(s_{T_i})\right)\right]
\end{equation}
 where $c_\theta(s_{T_i})$ is the estimated value of $s_{T_i}$
 produced by the critic for $\pi_\theta$. To further reduce variance, we follow \cite{schulman_iclr16}
 and estimate $Q(s_{T_i},g_i) \approx R_\theta(s_{T_{i}}) + \gamma c_\theta(s_{T_{i+1}})
 $ such that $Q(s_{T_i}, g_i) - c_\theta(s_{T_i})$ computes a
  generalized advantage estimator (GAE).
 Similarly, each \subpolicy $\pi_{\phi_g}$ is updated according to the
 gradient
 \begin{equation}
 \nabla_{\phi_g} J(\pi_{\phi_g}) = \mathbb{E} \left[ \nabla_{\phi_g} \log(\pi_{\phi_g}(a_i | s_i)) \left(Q(s_i, a_i) - c_{\phi_g}(s_i)\right)\right].
\end{equation}
 Recall from \secref{sec:policy} that these critics share parameters with their
 corresponding policy networks such that subgoals with a common task
 also share a critic.
We train each policy network independently using A3C~\cite{mnih_icml16} with
GAE \cite{schulman_iclr16} with 8 threads across 4 GPUs. After independent
reinforcement fine-tuning of the \subpolicies, we train the \master policy further using
the trained \subpolicies rather than expert \subgoal trajectories.

\textbf{Initial states and curriculum.} Rather than spawn agents at
    fixed distances from target, from where accomplishing the subgoal may
    be arbitrarily difficult,
    we sample locations along expert trajectories for each question or subgoal.
    This ensures that even early in training, policies are likely to have a
    mix of positive and negative reward episodes. At the beginning of training,
    all points along the trajectory are equally likely; however, as
    training progresses and success rate improves, we reduce the
    likelihood of sampling points nearer to the goal.
    This is implemented as a multiplier $\alpha$ on available states
    $[s_0, s_1, ..., s_{\alpha T}]$, initialized to $1.0$ and
    scaled by $0.9$ whenever success rate crosses a $40\%$ threshold.

\vspace{-10pt}
\section{Experiments and Results}
\label{sec:results}
\vspace{-5pt}

\begin{figure*}[t!]
    \begin{subfigure}{0.23\textwidth}
        \centering
        \caption{\small\texttt{Exit-room}}
        \label{fig:exit_room}
        \includegraphics[width=1.0\textwidth]{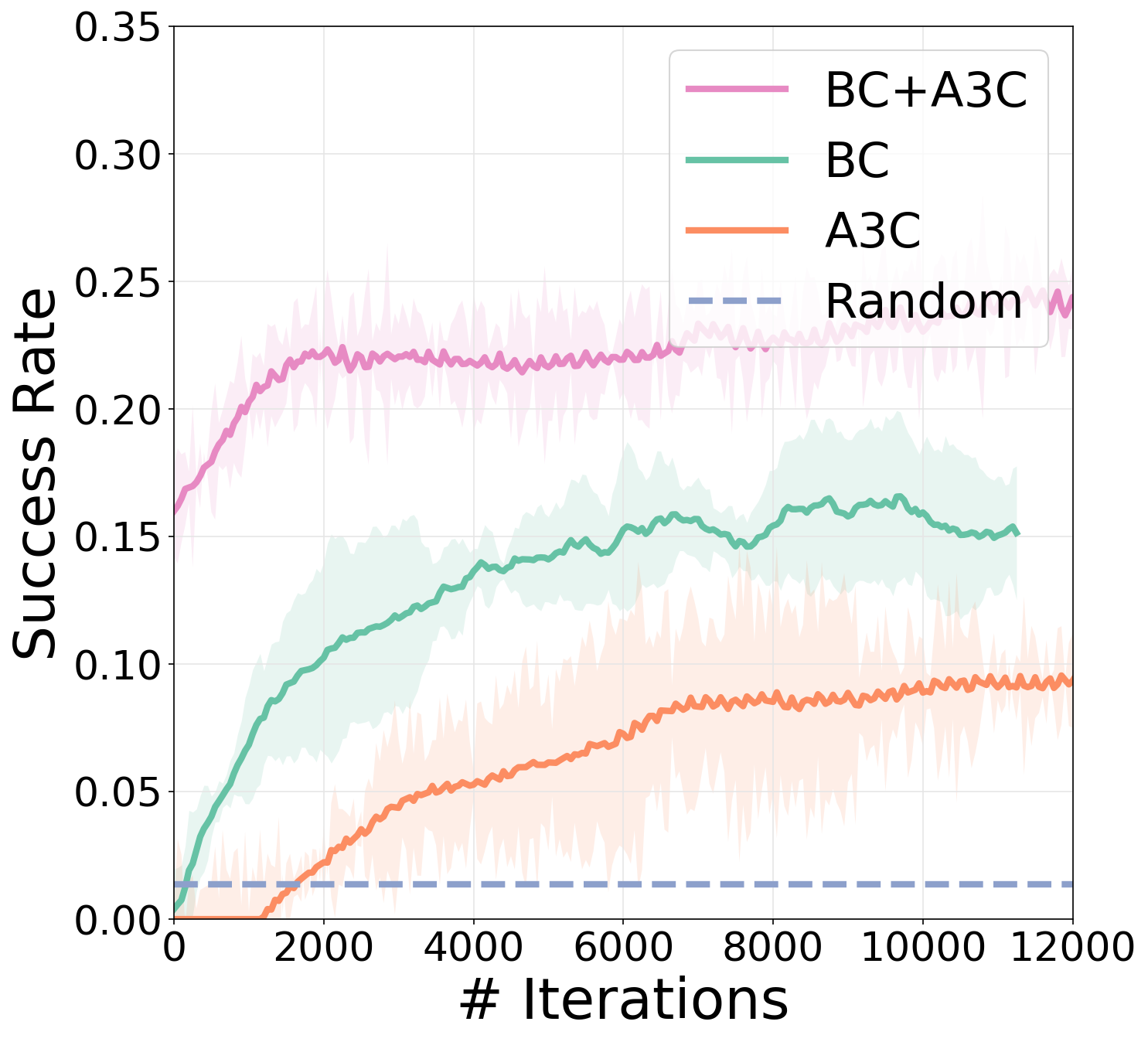}\\[5pt]
    \end{subfigure}
    \begin{subfigure}{0.23\textwidth}
        \centering
        \caption{\small\texttt{Find-room}}
        \label{fig:find_room}
        \includegraphics[width=1.0\textwidth]{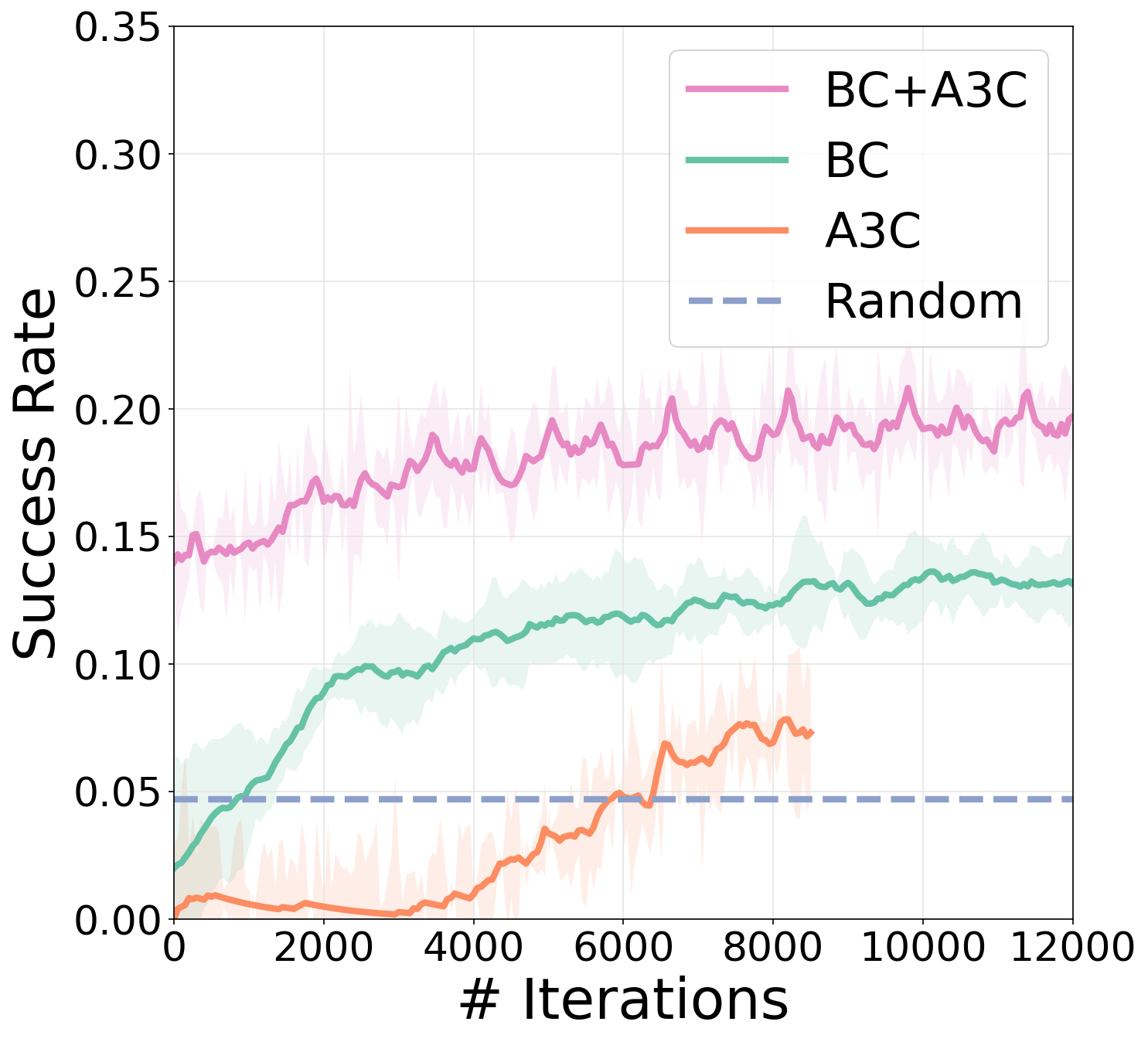}\\[5pt]
    \end{subfigure}
    \begin{subfigure}{0.23\textwidth}
        \centering
        \caption{\small\texttt{Find-object}}
        \label{fig:find_obj}
        \includegraphics[width=1.0\textwidth]{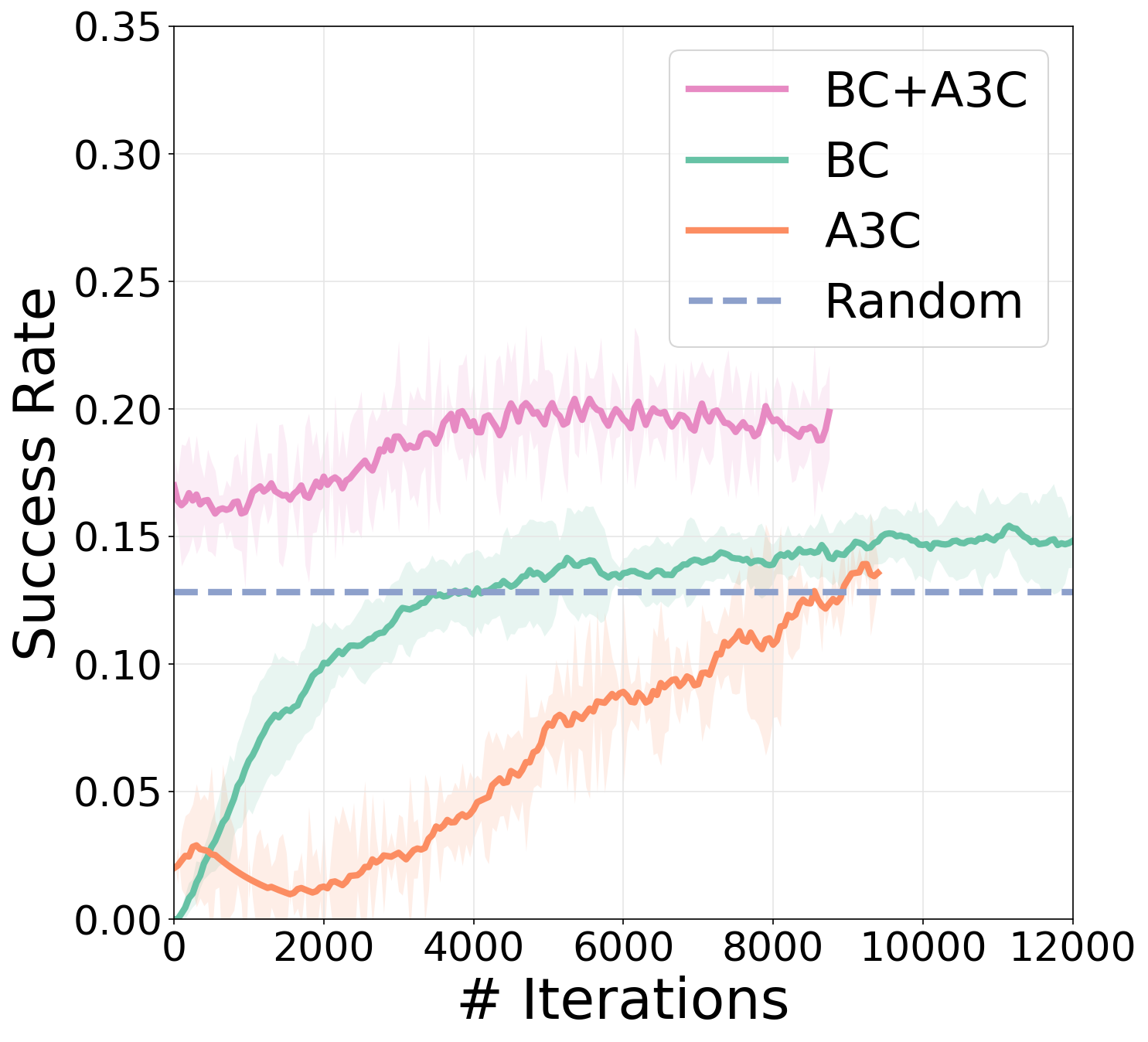}\\[5pt]
    \end{subfigure}
    \begin{subfigure}{0.29\textwidth}
        \centering
        \caption{{\small Loss curves for master policy}}
        \label{fig:master_loss}
        \includegraphics[width=0.75\textwidth]{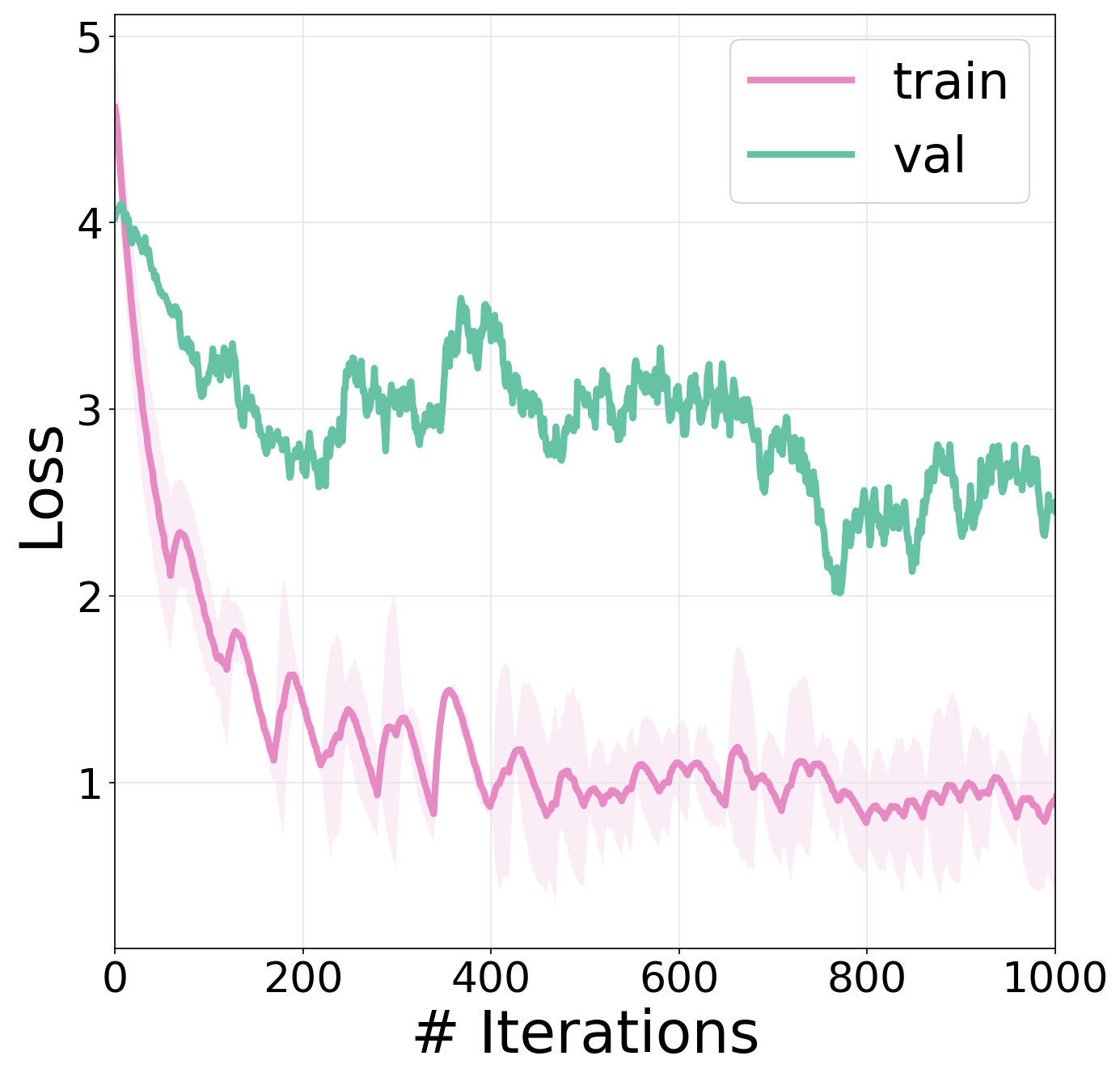}\\[5pt]
    \end{subfigure}
    \\[2pt]
    \caption{(a,b,c) Success rate over training iterations for each \subpolicy task using behavior cloning (BC),
    reinforcement learning from scratch (A3C), and reinforcement finetuning after behavior cloning (BC+A3C) training regimes.
    We find BC+A3C significantly outperforms either BC or A3C alone. Each of these is averaged over 5 runs.
    (d) Losses for master policy during behavior cloning \ie assuming access to perfect \subpolicies.}
    \label{fig:sub_perf}
\end{figure*}

\textbf{Dataset.} We benchmark performance on the EQA v1 dataset~\cite{embodiedqa}, which
contains ${\sim}9,000$ questions in $774$ environments --
split into $7129 (648)$ / $853 (68)$ / $905 (58)$ questions (environments)
for training/validation/testing respectively\footnote{Note that
the size of the publicly available dataset on {\small\url{embodiedqa.org/data}} is
larger than the one reported in the original
version of the paper due to changes in labels for color questions.}.
These splits have no overlapping environments between them,
thus strictly checking for generalization to novel environments.
We follow the same splits.

\textbf{Evaluating \subpolicies.} We begin by evaluating the performance of
 each \subpolicy with regard to its specialized task. For clarity, we break
 results down by \subgoal task rather than for each task-argument combination.
 We compare \subpolicies
 trained with behavior cloning (\textbf{BC}), reinforcement learning from
 scratch (\textbf{A3C}), and reinforcement fine-tuning after behavior
 cloning (\textbf{BC+A3C}). We also compare to a \textbf{random}
 agent that uniformly samples actions including {\small\texttt{stop}}
 to put our results in context. For each, we report the success rate
 (as defined in \reftab{tab:heuristics}) on the \eqads v1 validation
 set which consists of $68$ novel environments unseen during training.
We spawn \subpolicies at randomly selected suitable rooms
(\ie~{\small\texttt{Find-object[\emph{sofa}]}} will only
be executed in a room with a sofa) and allow them to execute for
a maximum episode length of 50 steps or until they terminate.

 \figref{fig:sub_perf} shows success rates for the different
 subgoal tasks over the course of training. We observe that:
 \begin{compactenum}[\hspace{5pt}-]

 \item \textbf{Behavior cloning (BC) is more sample-efficient than A3C from scratch.}
 Sub-policies trained using BC improve significantly faster than A3C for all tasks, and achieve
 higher success rates for {\small\texttt{Exit-room}} and {\small\texttt{Find-room}}. Interestingly,
 this performance gap is larger for tasks where a random policy does \emph{worse} -- implying that BC
 helps more as task complexity increases.

 \item \textbf{Reinforcement Fine-Tuning with A3C greatly improves over BC training alone.}
 Initializing A3C with a policy trained via behavior cloning results in a model that significantly
 outperforms either approach on its own, nearly doubling the success rate of behavior cloning for
 some tasks. Intuitively, mimicking expert trajectories in behavior cloning provides dense feedback
 for agents about how to navigate the world; however, agents never have to face the consequences
 of erroneous actions \eg recovering from collisions with objects -- a weakness that A3C fine-tuning addresses.
 \end{compactenum}

\begin{table*}[t]
\setlength\tabcolsep{3pt}
\renewcommand{\arraystretch}{1.2}
\resizebox{\textwidth}{!}{
\begin{tabular}{l l l c c c l c c c l c c c l c c c }
& & \multicolumn{12}{c}{\textbf{Navigation}} &~~~ & \multicolumn{3}{c}{\textbf{QA}}\\
\cmidrule{4-14}\cmidrule{16-18}
&&& \multicolumn{3}{c}{$\mathbf{d_0}$ {\scriptsize(For reference)}} & & \multicolumn{3}{c}{$\mathbf{d_T}$ {\scriptsize(Lower is better)}} & & \multicolumn{3}{c}{$\mathbf{d_\Delta}$ {\scriptsize(Higher is better)}} & & \multicolumn{3}{c}{$\mathbf{accuracy}$ {\scriptsize(Higher is better)}}\\
& && \scriptsize$T_{-10}$ & \scriptsize$T_{-30}$ & \scriptsize$T_{-50}$ &
& \scriptsize$T_{-10}$ & \scriptsize$T_{-30}$ & \scriptsize$T_{-50}$ &
 & \scriptsize$T_{-10}$ & \scriptsize$T_{-30}$ & \scriptsize$T_{-50}$ &
 & \scriptsize$T_{-10}$ & \scriptsize$T_{-30}$ & \scriptsize$T_{-50}$ \\
 \toprule
&PACMAN (BC)~\cite{embodiedqa} && 1.15 & 4.87 & 9.64 && 1.19 & 4.25 & 8.12 && -0.04  & 0.62 & 1.52 && 48.48\% & 40.59\% & 39.87\%\\
&PACMAN (BC+REINFORCE)~\cite{embodiedqa} && 1.15 & 4.87 & 9.64 && \textbf{1.05} & 4.22 & 8.13 && \textbf{0.10} & 0.65 & 1.51 && 50.21\% & 42.26\% & 40.76\% \\
\cmidrule{2-18}
& NMC (BC) && 1.15 & 4.87 & 9.64 && 1.44 & 4.14 & 8.43 && -0.29 & 0.73 & 1.21 && 43.14\% & 41.96\% & 38.74\% \\
& NMC (BC+A3C) && 1.15 & 4.87 & 9.64 && 1.06 & \textbf{3.72} & \textbf{7.94} && 0.09 & \textbf{1.15} & \textbf{1.70} && \textbf{53.58\%} & \textbf{46.21\%} & \textbf{44.32\%} \\
\bottomrule
\end{tabular}}\\[3pt]
\caption{Evaluation of \eqa agents on navigation and answering metrics for the \eqads v1 test set.}
\label{tab:results}
\end{table*}

\textbf{Evalating master policy.}
Next, we evaluate how well the master policy performs during independent behavior
cloning on expert trajectories \ie assuming perfect \subpolicies, as specified in
Eq.~\refeq{eq:master_cloning}. Even though there is no overlap between training and
validation environments, the master policy is able to generalize reasonably
and gets $\sim 48\%$ intersection-over-union (IoU) with ground truth subgoal sequences
on the validation set. Note that a sequence of sub-goals that is different from
the one corresponding to the shortest path may still be successful at navigating
to the target object and answering the question correctly. In that sense, IoU against
ground truth subgoal sequences is a strict metric. \reffig{fig:master_loss} shows the training and validation
cross-entropy loss curves for the master policy.

\textbf{Evalating NMC.}
Finally, we put together the master and \subpolicies and evaluate
navigation and question answering performance on \eqa.
We compare against the PACMAN model proposed in~\cite{embodiedqa}.
For accurate comparison, both PACMAN and NMC use the same publicly available
and frozen pretrained CNN\footnote{{\small\url{github.com/facebookresearch/EmbodiedQA}}},
and the same visual question answering model -- pretrained to predict answers
from last $5$ observations of expert trajectories, following~\cite{embodiedqa}.
Agents are evaluated by spawning $10$, $30$, or $50$
primitive actions away from target, which corresponds to distances of $1.15$, $4.87$, and $9.64$
meters from target respectively, denoted by $\mathbf{d_0}$ in \reftab{tab:results}.
When allowed to run free from this spawn location,
$\mathbf{d_T}$ measures final distance to target (how far is the agent
from the goal at termination), and $\mathbf{d_\Delta} = \mathbf{d_T} - \mathbf{d_0}$ evaluates change in
distance to target (how much progress does the agent make over the course of its navigation).
Answering performance is measured by $\mathbf{accuracy}$ (\ie did the predicted answer match ground-truth).
Note that \cite{embodiedqa} report a number of additional metrics (percentage of times the agent stops,
retrieval evaluation of answers, \etc).
Accuracies for PACMAN are obtained by running the publicly available codebase accompanying~\cite{embodiedqa}, 
and numbers are different than those reported in the original version of~\cite{embodiedqa} due to changes in the dataset$^1$.

As shown in \reftab{tab:results}, we evaluate two versions of our model -- 1) NMC (BC) naively combines master and
\subpolicies without A3C finetuning at any level of hierarchy, and 2) NMC (BC+A3C)
is our final model where each stage is trained with BC+A3C, as described in \refsec{sec:approach}.
As expected, NMC (BC) performs worse than NMC (BC + A3C), evident in worse navigation
$\mathbf{d_T}$, $\mathbf{d_\Delta}$ and answering $\mathbf{accuracy}$.
PACMAN (BC) and NMC (BC) go through the same training regime,
and there are no clear trends as to which is better -- PACMAN (BC) has better $\mathbf{d_{\Delta}}$
and answering $\mathbf{accuracy}$ at $T_{-10}$ and $T_{-50}$, but worse at $T_{-30}$.
No A3C finetuning makes it hard for \subpolicies to recover from erroneous primitive actions,
and for master policy to adapt to \subpolicies. A3C finetuning significantly boosts
performance, \ie NMC (BC + A3C) outperforms PACMAN with higher $\mathbf{d_\Delta}$ (makes more progress towards target),
lower $\mathbf{d_T}$ (terminates closer to target), and higher answering $\mathbf{accuracy}$.
This gain primarily comes from the choice of \subgoals and the
master policy's ability to explore over this space of \subgoals instead of primitive
actions (as in PACMAN), enabling the master policy to operate over longer
time horizons, critical for sparse reward settings as in \eqa.



\section{Conclusion}
\label{sec:conc}

We introduced Neural Modular Controller (NMC), a hierarchical policy for \eqa consisting of
a master policy that proposes a sequence of semantic subgoals from question
(\eg \myquote{What color is the sofa in the living room?} $\rightarrow$
~{\small\texttt{Find-room[living]}}, {\small\texttt{Find-object[sofa]}}, {\small\texttt{Answer}}),
and specialized \subpolicies for executing each of these tasks. The master and \subpolicies are
trained using a combination of behavior cloning and reinforcement learning, which is
dramatically more sample-efficient than each individual training regime. In particular,
behavior cloning provides dense feedback for how to navigate, and reinforcement learning
enables policies to deal with consequences of their actions, and recover from errors.
The efficacy of our proposed model is demonstrated on the EQA v1 dataset~\cite{embodiedqa}, where NMC
outperforms prior work both in navigation and question answering.

\subsubsection*{Acknowledgments}

This work was supported in part by NSF, AFRL, DARPA, Siemens, Google, Amazon,
ONR YIPs and ONR Grants N00014-16-1-$\{$2713,2793$\}$. The views and conclusions
contained herein are those of the authors and should not be interpreted as
necessarily representing the official policies or endorsements, either expressed
or implied, of the U.S. Government, or any sponsor.

%

\pagebreak

{
\small
\bibliography{strings,main}
}

\end{document}